\pdfoutput=1

\documentclass[%
 preprint,
 amsmath,amssymb,
 aps,
prl
]{revtex4-2}

\usepackage[english]{babel}
\usepackage{graphicx}% Include figure files
\usepackage{dcolumn}% Align table columns on decimal point
\usepackage{bm}% bold math
\usepackage{nicefrac}
\usepackage{hyperref}% add hypertext capabilities
\usepackage{soul}
\usepackage{tikz}
\usepackage{comment}

\newcommand{\mat}[1]{\bm{\mathrm{#1}}}

\renewcommand{\vec}[1]{\bm{\mathrm{#1}}}
\newcommand{\neuralfun}{\vec{f}}
\newcommand{\force}{\vec{F}}
\newcommand{\energy}{E}
\newcommand{\mass}{\mat{M}}
\newcommand{\y}{\vec{y}}
\DeclareMathOperator{\dd}{d}
\newcommand{\vtheta}{\vec{\theta}}
\newcommand{\veta}{\vec{\eta}}
\newcommand{\vphi}{\vec{\phi}}
\newcommand{\tmax}{T}
\newcommand{\norm}[1]{\lVert\,#1\,\rVert}

\graphicspath{{./Figures/}}

\begin{document}

\preprint{APS/123-QED}

\title{Meta-learning generalizable dynamics from trajectories}% Force line breaks with \\


\author{Qiaofeng Li$^{1,2,3}$}
\author{Tianyi Wang$^{2}$}
\author{Vwani Roychowdhury$^{2,*}$}
\author{M.K. Jawed$^{1,}$}
\email{V.R.: vwani@ee.ucla.edu, M.K.J.: khalidjm@seas.ucla.edu}
\affiliation{\footnotesize 
$^1$Dept.\ of Mechanical and Aerospace Engineering, University of California, Los Angeles, CA 90095, USA \\
$^2$Dept.\ of Electrical and Computer Engineering, University of California, Los Angeles, CA 90095, USA \\
$^3$Dept. \ of Mechanical Engineering, Massachusetts Institute of Technology, Cambridge, MA 02139, USA
}

%\begin{comment}
\begin{abstract}
We present the interpretable meta neural ordinary differential equation (iMODE) method to rapidly learn generalizable (i.e. not parameter-specific) dynamics from trajectories of multiple dynamical systems that vary in their physical parameters. The iMODE method learns meta-knowledge, the functional variations of the force field of dynamical system instances without knowing the physical parameters, by adopting a bi-level optimization framework: an outer level capturing the common force field form among studied dynamical system instances and an inner level adapting to individual system instances. 
\emph{A priori} physical knowledge can be conveniently embedded in the neural network architecture as inductive bias, such as conservative force field and Euclidean symmetry. With the learned meta-knowledge, iMODE can model an unseen system within seconds, and inversely reveal knowledge on the physical parameters of a system, or as a \emph{Neural Gauge} to ``measure'' the physical parameters of an unseen system with observed trajectories. We test the validity of the iMODE method on bistable, double pendulum, Van der Pol, Slinky, and reaction-diffusion systems.
\end{abstract}

\pacs{Valid PACS appear here}

\maketitle

%\tableofcontents

%\note{We are representing a Slinky using it's center line. We have a beam bending system that does not follow Euler-Bernoulli beam bending. This is an unknown system; however, ``secretly" we know that this unknown beam is just the centerline of a Slinky. This is a puzzle that no one will be able to solve (i.e., looking at a beam, someone cannot possibly infer that it is representing a Slinky). We can think about presenting this a testcase of an ``unknown" system.}

Building predictive models of dynamical systems is a central challenge across diverse disciplines of science and engineering. Traditionally, this has been achieved by first manually deriving the governing equations with carefully chosen state variables and then fitting the undetermined physical parameters using observed data, e.g.,~\cite{Sprakel2011Stress,Jawed2015Untangling,Alert2022Cellular}. In order to avoid the painstaking formulation of analytical equations, researchers have recently leveraged advances in machine learning and the data-fitting power of neural networks (NNs) to make the modeling process both automatic and more expressive~\cite{Karniadakis2021Physics}. This is achieved by either adopting the conventional physics-based approach as a starting point and then replacing various components with data-driven modules \cite{raissiDeepHiddenPhysics2018,Chen2018Neural}, or directly learning discrete dynamics using autoregressive models from high-dimensional observations \cite{Steven2016Discovering,championDatadrivenDiscoveryCoordinates2019,chenAutomatedDiscoveryFundamental2022}. These works, while promising, need to fit dedicated models separately for different system instances with different parameters, which limits a model's applicability to one specific instance.

% Note: it might not be clear what meta-knowledge means and in what sense the generalization ability is used. More explanation might endanger the page limit.
In this letter, our goal is to learn meta-knowledge, the form of dynamics that is unrestricted to specific physical parameters or initial/boundary conditions, on dynamical systems to reveal physical insights~\cite{Iten2020Discovering,Liu2021Machine,Liu2022Machine} and to significantly improve the generalization ability of data-driven models. Specifically, we learn the shared dynamics form from the trajectories generated by a series of dynamical system instances in spite of their diversified behaviors in data, \emph{without knowing the system parameters}. This separates our work from Refs.~\cite{leeParameterizedNeuralOrdinary2021,desaiOneShotTransferLearning2022} and Neural Operators \cite{Li2020Neural,liFourierNeuralOperator2020,Lu2021Learning,Wang2021Learning}, in which true parameters should be provided. This goal aligns with that of multi-task meta-learning \cite{wangBridgingMultiTaskLearning2021}, which aims to leverage the similarities between different tasks to enable better generalization and efficient adaptation to unseen tasks. %In particular, a line of works is termed gradient-based meta-learning (GBML) \cite{Finn2017Model,Nichol2018First,Finn2019Online,rajeswaranMetaLearningImplicitGradients2019,raghuRapidLearningFeature2019}, which, given a set of tasks and a neural network model, tries to find an initialization of the model such that the adaptation to a task takes only a small number of gradient descent steps while resulting in good generalization on the task's test data.

We propose an efficient and interpretable method to model a family of
dynamical systems using their observed trajectories, by combining gradient-based meta-learning (GBML) \cite{Finn2017Model,Nichol2018First,Finn2019Online,rajeswaranMetaLearningImplicitGradients2019,raghuRapidLearningFeature2019} with neural ordinary
differential equations (NODE)
\cite{Chen2018Neural,Chen2021Learning,Li2022Rapidly}. In recognizing that the systems have shared dynamics form
and varying physical parameters, we separate the model parameters into two parts: the \emph{shared parameters} that capture the shared form of dynamics, i.e. the meta-knowledge, and the \emph{adaptation parameters} that account for variations across
system instances. The method generalizes well on unseen systems from the same family,
and the adaptation parameters show good interpretability. The intrinsic
dimension of the varying system parameters can be estimated by analyzing the
adaptation parameters. Given ground truth of the system parameters, simple correspondence can be established between the adaptation parameters and actual physical parameters through diffeomorphism, which can be utilized as a ``\emph{Neural Gauge}''
to measure properties of new systems through observed trajectories. We name
our method interpretable meta neural ODE (iMODE).

In a general autonomous second-order system, the state of the system $\y$ contains the position (generalized coordinates) $\mathbf{x}$ and the velocity $\dot{\mathbf{x}}$. The dynamics of the second-order system is expressed by
\begin{equation}
    \dot{\y}
    =
    \begin{bmatrix}
    \dot{\mathbf{x}} \\
    \ddot{\mathbf{x}}
    \end{bmatrix}
    =
    \begin{bmatrix}
    \dot{\mathbf{x}} \\
    \mass^{-1}\force_{\vec{\phi}}(\y)
    \end{bmatrix}
    \text{, where }
    \y = \begin{bmatrix}
    \mathbf{x}\\
    \dot{\mathbf{x}}
    \end{bmatrix}
    % =
    % \begin{bmatrix}
    % \dot{\y} \\
    % \mass^{-1} \neuralfun_{\boldsymbol{\theta}}(\mathbf{x};\boldsymbol{\eta})
    % \end{bmatrix}
    \label{Eq:NN}
\end{equation}
where $\force_{\vphi}$ is the force vector containing all the internal and
external forces, and $\mass$ is the mass matrix. With a set of physical
parameters $\vphi$, the force function $\force(\cdot)$ dictates the
dynamics of the system, which determines a unique trajectory $\y(t)$ given an
initial condition $\y(t_0)$.
In the remainder of the letter, without loss of generality, mass is normalized to an identity matrix, i.e., $\mass=\mathbf{I}$.

Trajectories are collected from multiple system instances into a dataset $\mathcal{D}$.
Consider $N_\mathrm{s}$ instances that share the dynamics form $\force_{\vphi}(\cdot)$, but have distinct physical parameters, $\{\vec{\phi}_{1}, \ldots, \vec{\phi}_{N_\mathrm{s}}\}$ respectively. 
From each system instance, $N_\mathrm{tr}$ trajectories are observed, each containing observations across $T$ time steps. In summary, $\mathcal{D} = \left\{\{\y_{i,j}(t_k)\}_{k=0}^{\tmax} \vert i=1,\ldots,N_\mathrm{s},\; j=1,\ldots,N_\mathrm{tr}\right\}$.
The data-driven model is trained on $\mathcal{D}$, knowing which trajectories are from the same system instance (i.e. given both the index $i$ and $j$ of trajectories), but is not given the knowledge of $\{\vphi_i\}_{i=1}^{N_\mathrm{s}}$. 
% the following sentence is added, feel free to delete if space doesn't approve
Take the pendulum system as an example. An instance is a pendulum with a specific arm length (since the inertia is normalized), therefore $\vphi$ includes only the arm length. A trajectory contains the location and speed of the pendulum during a time period. 

\begin{figure}[t]
    \centering
    \includegraphics[width=0.49\textwidth]{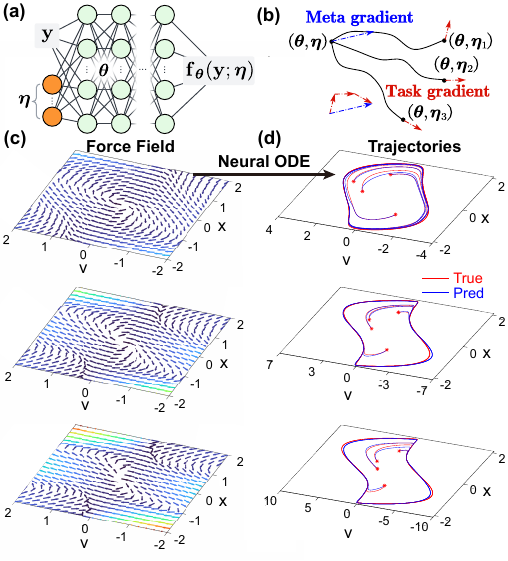}
    \caption{
    (color online). (a) In iMODE, a neural module $\force_{\vtheta}$ parameterized by $\vtheta$ takes the concatenation of system state $\y$ and the adaptation parameters $\boldsymbol{\eta}$ and generates the estimated force as output. (b) The bi-level iteration process in the iMODE method. The NN weights $\boldsymbol{\theta}$ are shared across system instances while $\boldsymbol{\eta}$ is adapted for each instance. The meta gradient w.r.t. $\boldsymbol{\theta}$ aggregates the gradients evaluated with instance-adapted $\boldsymbol{\eta}$.
    % (c) The learned evolution of the force field of the Van der Pol system with increasing $\epsilon$ from top down.
    % (d) The fast adapted force field can be used to predict system trajectories for unseen initial conditions with Neural ODE.
    (c) Examples of estimated force field $\neuralfun_{\vtheta}(\cdot;\veta)$
    for Van der Pol system instances that differ in their $\epsilon$ parameter
    (in ascending order from top to bottom). % $\veta$ is adapted to the system instances respectively to fit observed trajectories, and as $\veta$ changes, the corresponding $\neuralfun_{\vtheta}(\cdot;\veta)$ morphs in a regularized fashion. 
    The estimation quality is further evaluated through the trajectories generated by the fields as shown in (d).
    (d) The estimated force field can be used to predict system trajectories for
    unseen initial conditions through integration (Eq.~\eqref{Eq:yhat}). The
    signature limit cycles of Van der Pol systems are faithfully reproduced.}
    \label{fig:Figure1}
\end{figure}

In our framework, a neural network $\neuralfun_{\vtheta}(\y;\veta)$ (Fig.~\ref{fig:Figure1}(a).
% \revise{DenseNet-like fully connected networks are used as backbone architecture for tasks throughout this work;}
See Supplemental Material (SM) \cite{Supplement} for detailed description) replaces $\force_{\vphi}(\y)$ in Eq.~\eqref{Eq:NN} to approximate the observed trajectories, where $\boldsymbol{\eta}$ is adapted to each system instance such that with a certain $\veta_i$, $\neuralfun_{\vtheta}(\y; \veta_i)$ approximates the force function of the $i$th system instance $\force_{\vphi_i}(\y)$.
%\note{We will never get mass/gravity/length; but will get something similar that is represented by $\eta$.}
%\note{What is the difference between $\theta$ and $\eta$.}
After training, $\veta$ becomes a proxy for the physical parameters $\vphi$.
$\vtheta$ is the model parameters that capture the functional form of dynamics shared across system instances.
The predicted trajectory starting from an initial condition $\y_0$ is given by integration (the 5th-order Dormand-Prince-Shampine solver is used throughout this letter to compute integrals)
\begin{equation}
    \widehat{\y}(t, \y_0, \vtheta, \veta) = \y_0 + \int_{t_0}^t \neuralfun_{\vtheta}\left(\widehat{\y}(\tau); \veta\right) \dd \tau
    \label{Eq:yhat}
\end{equation}
% (the 5th-order Dormand-Prince-Shampine algorithm is used throughout this paper)

For brevity, we denote the trajectory $\y_{i,j}(t)$ as $\y_{i,j}$, the
corresponding prediction $\widehat{\y}(t, \y_{i,j}(t_0), \vtheta, \veta)$ as
$\widehat{\y}_{i,j}(\vtheta, \veta)$, and use $\norm{\y_{i,j} -
\widehat{\y}_{i,j}(\vtheta, \veta)}^2$ to denote
$\sum_{k=0}^{\tmax}\left(\y_{i,j}(t_k) - \widehat{\y}(t_k, \y_{i,j}(t_0),
\vtheta, \veta)\right)^2$, the squared difference between $\y_{i,j}$ and
$\widehat{\y}_{i,j}(\vtheta, \veta)$ across all time steps.

The goal of the modeling is formulated as a bi-level optimization (Fig.~\ref{fig:Figure1}(b)),% and $\veta$:
% \begin{align}
%         \textrm{outer}: \min_{\vtheta} & \quad \widetilde{\mathcal{L}}(\vtheta) = \frac{1}{N_\mathrm{s}} \sum_{i=1}^{N_\mathrm{s}} \mathcal{L}_i(\vtheta, \veta^{(m)}_i) \\
%         \text{where}          & \quad \mathcal{L}_i(\vtheta, \vec{\zeta}) = \frac{1}{N_\mathrm{tr} \tmax} \sum_{j=1}^{N_\mathrm{tr}} \norm{\y_{i,j} - \widehat{\y}_{i,j}(\vtheta, \vec{\zeta})}^2,\\
%                              \textrm{inner}: & \quad \veta^{(l+1)}_i = \veta^{(l)}_i - \alpha \nabla_{\veta} \mathcal{L}_i(\vtheta, \veta^{(l)}_i), \;\; \veta^{(0)}_i = \veta \label{eq:inner-opt}
% \end{align}
\begin{align}
        \textbf{outer:} & \quad  \min_{\vtheta} \ \widetilde{\mathcal{L}}(\vtheta) = \frac{1}{N_\mathrm{s}} \sum_{i=1}^{N_\mathrm{s}} \mathcal{L}_i(\vtheta, \veta^{(m)}_i) \text{, where} \\
        & \quad \ \mathcal{L}_i(\vtheta, \vec{\zeta}) = \frac{1}{N_\mathrm{tr} \tmax} \sum_{j=1}^{N_\mathrm{tr}} \norm{\y_{i,j} - \widehat{\y}_{i,j}(\vtheta, \vec{\zeta})}^2,\\
        \textbf{inner:}& \quad \veta^{(l+1)}_i = \veta^{(l)}_i - \alpha \nabla_{\veta} \mathcal{L}_i(\vtheta, \veta^{(l)}_i), \;\; \veta^{(0)}_i = \veta \label{eq:inner-opt}
\end{align}
%\note{Possibly add ``INNER LEVEL" and ``OUTER LEVEL" as headings before equations 4 and 5.}
%\note{What type of NNs were used? We can give this information in the Supp. Materials and also comment on the generalizability (model agnostic).}
where the inner-level involves an $m$-step gradient descent adapting $\veta$ for each instance, while the outer-level finds the optimal initialization for $\vtheta$. $\alpha$ is the inner-level stepsize and $\veta_i^{(m)}$ is the adaptation parameters for the $i$th system instance after $m$ steps of adaptation. For short, we denote such $i$th adaptation result as $\veta_i$. Note that $\veta_i$ depends on both $\vtheta$ and $\veta$ as shown in Eq.~\eqref{eq:inner-opt}.
To avoid higher-order derivatives, we simplify such dependency following the first-order Model Agnostic Meta-Learning (first-order MAML) \cite{Finn2017Model} and use the outer-level step as % (also referred to as meta-optimization) 
% \begin{equation}
%     \begin{gathered}
%     % (\vtheta, \veta) \leftarrow (\vtheta, \veta) - \beta \sum_i \nabla_{(\vtheta, \veta)} \mathcal{L}_i({\vtheta}, \veta^{(m)}_i), \\
%     % \text{assuming } \nabla_{\vtheta} \veta^{(m)}_i = 0, \nabla_{\veta} \veta^{(m)}_i = \mat{I}
%     (\vtheta, \veta) \leftarrow (\vtheta, \veta) - \beta \sum_i \nabla_{(\vtheta, \veta)} \mathcal{L}_i({\vtheta}, \veta_i), \\
%     \text{assuming } \nabla_{\vtheta} \veta_i = 0, \nabla_{\veta} \veta_i = \mat{I}
%     \end{gathered}
% \end{equation}
\begin{equation}
    \vtheta \leftarrow \vtheta - \frac{\beta }{N_{\mathrm{s}}}\sum_i \nabla_{\vtheta} \mathcal{L}_i({\vtheta}, \veta_i),\; \text{\footnotesize (assuming that $\dfrac{\partial \veta_i}{\partial \vtheta} = \mathbf{0}$)}
\end{equation}
where $\beta$ is the outer-level stepsize.
At both the inner-level and outer-level, the gradient calculation for functions involving integrals is enabled by NODE \cite{Chen2018Neural,Chen2021Learning,Li2022Rapidly}.
% $\boldsymbol{\theta}$ is the common NN weights shared across different tasks. $\boldsymbol{\eta}$ is the latent input adapted for each parametric system.

As shown in Fig.~\ref{fig:Figure1}(c), $\neuralfun_{\vtheta}(\cdot ; \veta)$ specifies a force field that morphs as $\veta$ changes. Note that $m$ is normally quite small (e.g. 5), so given trajectories of a previously unseen system, $\veta$ can be efficiently updated with few gradient steps, adapting the NN to specify a force field explaining behaviors of the new system, which is one order-of-magnitude faster compared to training from scratch
% while training the NN to the same accuracy from scratch would require one order-of-magnitude higher number of gradient steps 
(Fig.~\ref{fig:Figure3}(a)). Trajectories with arbitrary initial conditions can be inferenced based on the force field (Fig.~\ref{fig:Figure1}(d)).

\begin{figure}[htbp]
    \centering
    \includegraphics[width=0.65\textwidth]{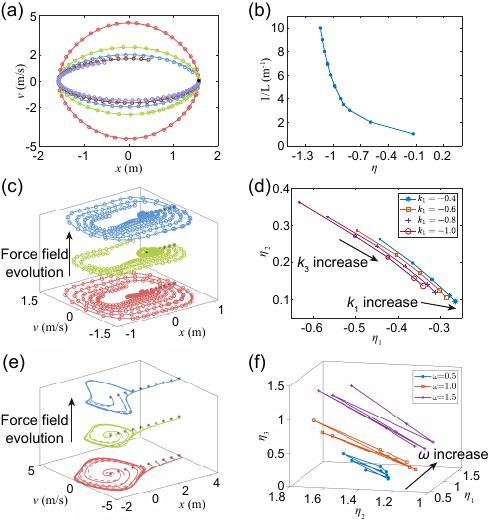}
    \caption{(color online). (a) The meta-learning results for the pendulum. The iMODE trajectory prediction (circles) with different arm lengths (different colors) match those of the ground truth (solid lines). (b) The learned $\veta$ is in good correlation with the effective stiffness of different pendulums ($1/L$). (c) The predicted trajectories (circles) match those of ground truth (solid lines) with different initial conditions (black stars) and different system parameters (different colors) for the bistable system. (d) Two principal axes can be identified from the latent space of the learned $\veta$, each regarding the variation of one physical parameter. (e) Similar to (c) but for the Van der Pol system. (f) The principal axis regarding to the variation of $\omega$ for the Van der Pol system.}
    \label{fig:Figure2}
\end{figure}

% Figure2 Results. Generally explain the results on different systems.
First we validate the modeling capability of the iMODE algorithm on 3 cases: oscillating pendulum, bistable oscillator, and Van der Pol system (see SM \cite{Supplement} for detailed description). The oscillating pendulum has 1 physical parameter, i.e. the arm length (rotational inertia normalized). Fig.~\ref{fig:Figure2}(a) shows that the predicted trajectories using task-adapted NNs match the ground truth of each system. Fig.~\ref{fig:Figure2}(b) shows that the learned $\boldsymbol{\eta}$ correlates well with the effective stiffness of the pendulum, i.e. $1/L$. Effectively $\boldsymbol{\eta}$ acts as a proxy of the true arm length and can be used to infer such parameters of unseen systems.

The bistable system has a potential energy function controlled by 2 parameters $k_1$ and $k_3$. Its potential energy has two local minima, or potential wells. When the initial conditions vary, the bistable system can oscillate intra-well or inter-well. Fig.~\ref{fig:Figure2}(c) shows that the task adapted trajectories ($m=5$) match the ground truth well. Fig.~\ref{fig:Figure2}(d) shows that the identified $\boldsymbol{\eta} \in \mathbb{R}^{2}$ has two principal axes, along which $k_1$ and $k_3$ increases. As mentioned, $\boldsymbol{\eta}$ is effectively a proxy for $k_1$ and $k_3$. Later we will show that the mapping from $\boldsymbol{\eta}$ to $\vphi=[k_1,k_3]$ can be constructed as a diffeomorphism with NODE.

The Van der Pol system has 3 physical parameters $\vphi= [\epsilon,\delta,\omega]$. It exhibits limit cycles due to the negative damping for small oscillation amplitudes. Fig.~\ref{fig:Figure2}(e) shows that the evolution of limit cycles due to the change of physical parameters is well predicted. 
%In this case, $\boldsymbol{\eta}$ is a proxy for $\epsilon$, $\delta$, and $\omega$. 
Three principal axes can be found for the identified $\boldsymbol{\eta}$. The one for $\omega$ is shown in Fig.~\ref{fig:Figure2}(f) (see SM \cite{Supplement} for the other two). Again, the mapping from $\boldsymbol{\eta}$ to $\vphi=[\epsilon,\delta,\omega]$ can be constructed as a diffeomorphism.

The fast adaptation of iMODE is demonstrated with the bistable systems in Fig.~\ref{fig:Figure3}(a). The iMODE is able to adjust the adaptation parameters in 5 steps to learn the dynamics of unseen system instances. Training the same network from scratch (random initialization) on the same test dataset requires much more epochs to achieve a comparable accuracy. When evaluated on trajectories with unseen initial conditions, the performance of iMODE-adapted models outperforms that of the model trained from scratch by several orders of magnitude, showing superior generalization ability with limited data
% 50 unseen bistable systems with randomly chosen $k_1$ and $k_3$
(see SM \cite{Supplement} for a more disparate comparison when data is scarce).

Second, we demonstrate the combination of the iMODE algorithm with certain physics priors for efficient modeling of more complicated systems. Since iMODE does not assume specific architecture of $\neuralfun_{\vtheta}$, a wide range of neural network architectures can be adopted to embed appropriate inductive biases. For example, in bistable and the following wall bouncing and Slinky systems, the assumption of conservative force is introduced, where the system dynamics is determined by a potential energy function. Accordingly we take a specific form for the neural force estimator
\begin{math}
    \label{Eq:Energy}
   \neuralfun_{\vtheta} (\mathbf{x};\boldsymbol{\eta}) = {\partial \energy_{\vtheta} (\mathbf{x};\boldsymbol{\eta})} / {\partial \mathbf{x}}
\end{math}.
That is, the NN first outputs an energy field and then induces the force field from the energy field (using auto-differentiation \cite{Paszke2019PyTorch}). In this way, iMODE enables the fast adaptation of not only the force field but also the potential energy field for the parametric systems. The learned potential energy functions are shown in Fig.~\ref{fig:Figure3}(b). The wall bouncing system has a potential energy well
% \begin{equation}
%     \energy(x) = \left\{
%     \begin{aligned}
%     &0, &|x| < w\\
%     &\infty, &|x| \geq w
%     \end{aligned}
%     \right.
% \end{equation}
% Note 
that is not a linear function of the well's \mbox{(half-)width} $w$ or the particle position $x$ (see SM \cite{Supplement}). However, iMODE is still able to approximate the discontinuous energy function. $\boldsymbol{\eta}$ correlates 
%shown to be 
well with the true width $w$, i.e., we can control the width of the potential energy well by tuning $\boldsymbol{\eta}$ (see SM \cite{Supplement}).

\begin{figure}[t]
    \centering
    \includegraphics[width=0.75\textwidth]{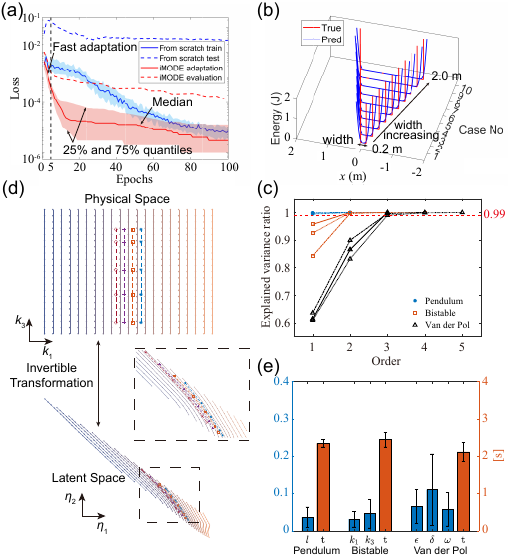}
    \caption{(color online). (a) Comparison of iMODE test adaptation v.s. training from scratch on (50) unseen bistable system instances with randomly chosen physical parameters. iMODE demonstrates fast adaptation and good generalization within the first 5 adaptation steps. (b) The true and learned potential energy functions for the wall bouncing system. The width of the potential well increases as the adaptation parameter increases. (c) The number of top PCA components that preserve a significant portion ($>$ 99\%) of the variance gives a good estimation of the dimension of true physical parameters. (d) The diffeomorphism constructed by NODE for the bistable system. It shows how a grid in the physical space is continuously deformed into the latent space of adaptation parameters. (e) The mean error and computation time of \emph{Neural Gauge} for 100 systems with randomly generated unseen parameters.}
    \label{fig:Figure3}
\end{figure}

The intrinsic dimension $d_{\vphi}$ of the physical parameters $\vphi$ can be estimated by applying Principal Component Analysis (PCA) to the collection of the $\veta$ vectors, each adapted to one of the system instances. Using an ``elbow'' method on the cumulative explained variance ratio curve of the PCA result, the number of the principal components that explain the most of the variance has a good correspondence with $d_{\phi}$, as long as $d_{\veta} \geq d_{\vphi}$, where $d_{\veta}$ is the dimension chosen for $\veta$.
% similar to \cite{zhengUnsupervisedLearningLatent2018a}
The PCA results on the pendulum, bistable system, and Van der Pol system are shown in Fig.~\ref{fig:Figure3}(c) (see SM \cite{Supplement} for the results of other systems). Taking the Van der Pol system as an example, $d_{\veta}$ is respectively $3$, $4$ or $5$ for the three curves with triangle markers. In all three cases, the first three principal components explain more than $99\%$ of the variance, and the ``elbow'' appears at $3$, which corresponds well with the fact that $d_{\vphi} = 3$ for the Van der Pol system.%, since $\vphi = \begin{bmatrix} \epsilon & \delta & \omega \end{bmatrix}$

% \fix{Figure 3(d) and (e). Neural Gauge} How to quickly determine the parameters of an unseen system from its trajectories.

\emph{Neural Gauge}: Without labels for the physical parameters, iMODE develops a latent space of adaptation parameters accounting for the variations in dynamics among system instances. Given the physical parameter labels of the system instances in the training data, a mapping between the space of the physical parameters and the latent space can be established so that the corresponding physical parameters can be estimated given any point in the latent space. iMODE therefore can be exploited as a ``Neural Gauge'' to identify the physical parameters of unseen system instances, and the establishment of such mappings can be seen as a calibration process. We propose to construct such mappings as diffeomorphism, which can be learned with a neural ODE ${\mathrm{d}\mathbf{z}(t)} / {\mathrm{d} t} = \vec{g}_{\boldsymbol{\xi}}(\mathbf{z})$, such that starting from a given point in the latent space, $\mathbf{z}(0) = \boldsymbol{\eta}_i$, the state $\vec{z}$ at $t=1$ gives the corresponding physical parameters, $\mathbf{z}(1) = \boldsymbol{\phi}_i$, $i=1,\ldots,N_s$.
For simplicity, the dimension of the latent space and that of the physical parameter space are assumed to match ($d_{\veta} = d_{\vphi}$); see SM \cite{Supplement} for more general treatment.
% \emph{Neural Gauge}: the iMODE algorithm can be further exploited for direct identification of the physical parameters of unseen cases, when the physical parameters of training cases are known (sparsely known since we only have less than 30 cases in the above cases). A diffeomorphism from the latent input space to the physical parameter space is constructed through training the following NN
% \begin{equation}
%     \begin{aligned}
        % \label{Eq:Diffeomorphism}
        % &\frac{\mathrm{d}\mathbf{z}(t)}{\mathrm{d} t} = \vec{g}_{\boldsymbol{\xi}}(\mathbf{z}), \text{ such that when } \mathbf{z}(0) = \boldsymbol{\eta}_i,\\
        % &\mathbf{z}(1) = \boldsymbol{\phi}_i, \; i=1,\dots,N_s % \\
         % \ (i=1,2,\dots,N_s)
%     \end{aligned}
% \end{equation}
$\vec{g}_{\boldsymbol{\xi}}$ is a NN whose weights are optimized by 
\begin{math}
    \boldsymbol{\xi} = \mathop{\arg\min}_{\boldsymbol{\xi}} \sum_i \Vert \mathbf{z}_i(1) - \boldsymbol{\phi}_i \Vert_2^2.
\end{math}

Figure~\eqref{fig:Figure3}(d) shows the learned diffeomorphism for the bistable system. The diffeomorphism establishes a bijection between the physical space and the latent space so that a grid in the physical parameter space can be continuously transformed into the adaptation parameter space (see SM \cite{Supplement}). The visualization highlights the advantages of diffeomorphism mapping: (1) The transformation is smooth so that the local geometric structure is preserved; (2) Invertible transformation allows better interpretation of the latent space compared to degenerating ones.

After constructing the diffeomorphism, we test the physical parameter identification performance on 100 randomly selected unseen instances (with random physical parameters). The identification error and time cost are shown in Fig.~\eqref{fig:Figure3}(e) for pendulum, bistable, and Van der Pol systems. The end-to-end identification starting from data-feeding normally takes around 2 seconds.% (the entire process of (1) taking in the trajectories of a new system, (2) adapting the adaptation parameters, and (3) estimating the physical parameters through the diffeomorphism).

% \begin{figure}[t]
%     \centering
%     \includegraphics[width=0.48\textwidth]{Figure4.pdf}
%     \caption{(color online). (a) The testing performance of the iMODE model on an unseen Slinky (of an unseen Young's modulus) with the same boundary condition as the training dataset. Top and bottom rows are ground truth and the iMODE model prediction at 0.28, 0.47, 0.65, 0.83 s (left to right). (b) The testing performance of the iMODE model on unseen initial and boundary conditions. Top and bottom rows are ground truth and the iMODE model prediction at 0.15, 0.32, 0.48, 0.65 s (left to right).}
%     \label{fig:Figure4}
% \end{figure}

% the original version with Figure 4 in the main draft
% We further demonstrate that iMODE applies to complex systems with two examples: a 40-cycle Slinky and a reaction-diffusion system described by the Kolmogorov-Petrovsky-Piskunov (KPP) equation (see SM \cite{Supplement} for the latter). In the Slinky case, we are able to embed Euclidean invariance for the energy field and induce equivariance for the force field. iMODE is able to learn from 4 Slinky cases (of Young's modulus 50, 60, 70, and 80 GPa, dropping under gravity from a horizontal initial configuration with both ends fixed) and then quickly generalize (with 2 adaptation steps) to an unseen Slinky (of Young's modulus 56 GPa) under unseen initial and boundary conditions, as shown in Fig.~\ref{fig:Figure4} (refer to \cite{Supplement,Li2022Rapidly} for details).
% the revised version with Slinky and reaction-diffusion system both moved to supplement

\emph{Complex systems}: We further demonstrate that iMODE applies to complex systems with two examples: a 40-cycle Slinky and a reaction-diffusion system described by the Kolmogorov-Petrovsky-Piskunov (KPP) equation. In the Slinky case, we embed Euclidean invariance for the energy field and induce equivariance for the force field. iMODE is able to learn from 4 Slinky cases (of Young's modulus 50, 60, 70, and 80 GPa, dropping under gravity from a horizontal initial configuration with both ends fixed) and then quickly generalize (with 2 adaptation steps) to an unseen Slinky (of Young's modulus 56 GPa) under unseen initial and boundary conditions. In the KPP equation case, iMODE is able to learn the reaction term with different reaction strength coefficients in 5 adaptation steps under Neumann boundary conditions and directly generalize to unseen Dirichlet boundary conditions. Refer to SM \cite{Supplement} for details.

% To solve the KPP equation, we discretize the solution domain [0,1] into 20 segments. So the the partial differential equation system (here $x$ denotes the space domain)
% \begin{equation}
% \frac{\partial u}{\partial t}=D \frac{\partial^2 u}{\partial x^2}+r u(1-u)
% \end{equation}
% is represented by an ordinary differential equation system containing 21 variables. The diffusion term is approximated by 2nd-order central difference and the diffusivity is assumed known. The meta-learning is performed to learn the reaction term with different reaction strength coefficients $r$. The training dataset contains 5 systems with $r=0.01,\ 0.02,\ 0.03,\ 0.04,\ 0.05$. The Neumann boundary condition $u^{\prime}(0)=u^{\prime}(1)=0$ ($\prime$ denotes derivative with respect to $x$) is used across the training dataset. The training results for $r=0.01,\ 0.05$ are shown in Fig.~\ref{fig:Figure4}(c). The iMODE NN is then adapted on the data from a unseen system with $r=0.034$. The resulted NN is directly applied to computation with unseen initial and boundary conditions (Dirichlet type $u(0)=u(1)=1$). The results of the latter are shown in Fig.~\ref{fig:Figure4}(d) and a good agreement is observed. This again validates the capability of the iMODE algorithm to fast adapt on unseen complex parametric systems and accurately predict on different initial and boundary conditions from those in the training dataset.

We have presented the iMODE method, i.e., interpretable meta NODE. As a major difference from existing NN-based methods, iMODE learns meta-knowledge on a family of dynamical systems, specifically the functional variation of the derivative (force) field. It constructs a parametrized functional form of the derivative field with a shared NN across system instances and latent adaptation parameters adapted for different instances. The NN and adaptation parameters are learned from the difference between the ground truth and the solution calculated by an appropriate ODE solver. 
We have validated with various examples the generalizability, interpretability, and fast adaptation ability of the iMODE method.
%(1) generalizability: iMODE learns the evolution of the derivative field. Once training is successfully completed, the resulting NN is able to make accurately predictions on unseen initial and boundary conditions. (2) interpretability: the latent input acts as a surrogate to the true physical parameters. With this feature, we can extract physical insights from experimental data of systems not fully understood. (3) fast adaptation: we have demonstrated that the task adaptation to an unseen system can be completed within a few adaptation epochs. 
iMODE opens a potential avenue for modeling and real-time control problems where the underlying systems are rapidly changing.

%\begin{acknowledgments}
%xxx
%\end{acknowledgments}

\appendix

\clearpage
\renewcommand{\thefigure}{S\arabic{figure}}
\setcounter{figure}{0}    

\begin{center}
{\Large Supplemental Information}
\end{center}

\section{Testing performance of the iMODE method}
\begin{figure}[htbp]
    \centering
    \includegraphics[width=0.95\textwidth]{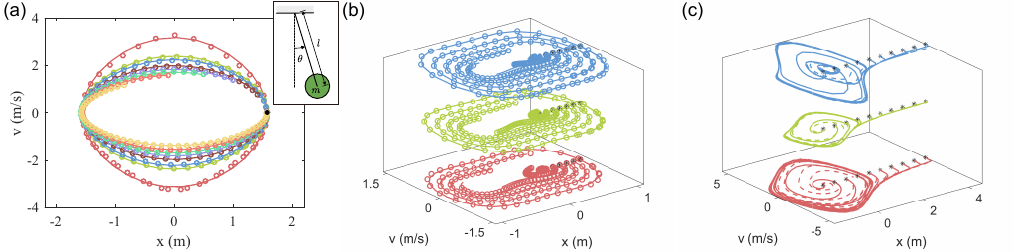}
    \caption{(color online). The testing performance of (a) the pendulum system, (b) the bistable system, (c) the Van der Pol system. The solid lines are ground truth. In (a) and (b) the circles, in (c) the dashed lines, are predictions of corresponding iMODE models. Different colors represent different parametric systems.}
    \label{fig:Testing}
\end{figure}
\subsection{Oscillating pendulum}
\begin{align}
    m l^2\ddot{\theta} + m g l \sin{(\theta)} = 0 \quad \text{s.t.} \quad \theta(0) = \theta_0, \ \dot{\theta}(0) = \dot{\theta}_0
\end{align}
The training dataset contains 5 system instances with $l=[1,3,5,7,9]$ m. A long trajectory of 10 s is generated for each instance with the initial position and velocity $\pi/2$ and 0 $s^{-1}$ and time marching stepsize 10 ms. During training, a batch of 20 trajectories of 1 s are pulled out randomly in each epoch. So essentially the iMODE training is seeing 1 s trajectories with differential initial conditions.

The learnt iMODE model is tested on 8 unseen system instances with $l=[2,3.5,4,5.1,6,6.9,8,10]$ m. The task adaptation is similarly done as the training, i.e., seeing a batch of 20 randomly pulled-out trajectories of 1 s for each testing system instance. The task adaptation only takes 5 steps. Then the learnt model for each instance is used to calculate a trajectory of 5 s given an initial condition, and compared with ground truth. The results are shown in Fig.~\ref{fig:Testing}(a). The solid lines (the ground truth) match well with the circles (prediction). 

\subsection{Bistable oscillator}
\begin{equation}
    \ddot{x} + k_1 x + k_3 x^3 = 0 \quad s.t. \quad x(0) = x_0, \ \dot{x}(0) = \dot{x}_0
\end{equation}
The training dataset contains 20 system instances, a mesh of $k_1=[-0.4,-0.6,-0.8,-1.0]$ and $k_3=[2.0,2.9,3.7,4.6,5.0]$. Trajectories of multiple initial conditions with stepsize 10 ms and time span 10 s are generated for each instance. During training, a batch of 100 randomly pulled-out trajectories of 1 s is used for each epoch. During testing, task adaptation takes 5 steps on previously unseen systems $[k_1,k_3] = [-0.5,3.1], [-0.7,4.2], [-0.5,4.7]$. The learnt models calculate trajectories of 5 s given an initial condition. The results are shown in Fig.~\ref{fig:Testing}(b).

\begin{figure}
    \centering
    \includegraphics[width=0.95\textwidth]{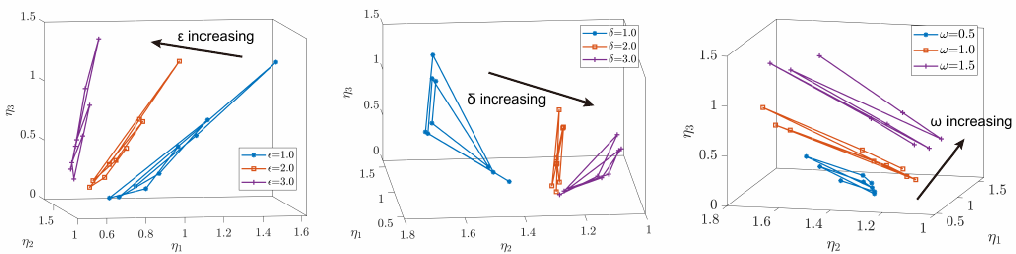}
    \caption{(color online). The three variation directions in the latent space for the physical parameters of the Van der Pol system $\epsilon$, $\delta$, and $\omega$.}
    \label{fig:VanderPol}
\end{figure}
\subsection{Van der Pol system}
\begin{equation}
    \ddot{x} - \epsilon \dot{x} (1-\delta x^2) + \omega^2 x = 0 \quad s.t. \quad x(0) = x_0, \ \dot{x}(0) = \dot{x}_0
\end{equation}
The training dataset contains 27 system instances, a mesh of $\epsilon=[1.0,2.0,3.0]$, $\delta=[1.0,2.0,3.0]$, and $\omega=[0.5,1.0,1.5]$. Trajectories of multiple initial conditions with stepsize 10 ms and time span 10 s are generated for each instance. During training, a batch of 100 randomly pulled-out trajectories of 1 s is used for each epoch. During testing, task adaptation takes 5 steps on previously unseen systems instances $[\epsilon,\delta,\omega] = [1.2,1.2,2.1], [1.2,1.8,1.4], [2.6,1.5,2.5]$. The learnt models calculate trajectories of 5 s given an initial condition. The results are shown in Fig.~\ref{fig:Testing}(c).

The three variation directions in the latent space $\boldsymbol{\eta}\in \mathbb{R}^3$ for $\epsilon$, $\delta$, and $\omega$ are shown in Fig.~\ref{fig:VanderPol}.

\section{Other systems}
\begin{figure}[htbp]
    \centering
    \includegraphics[width=0.85\textwidth]{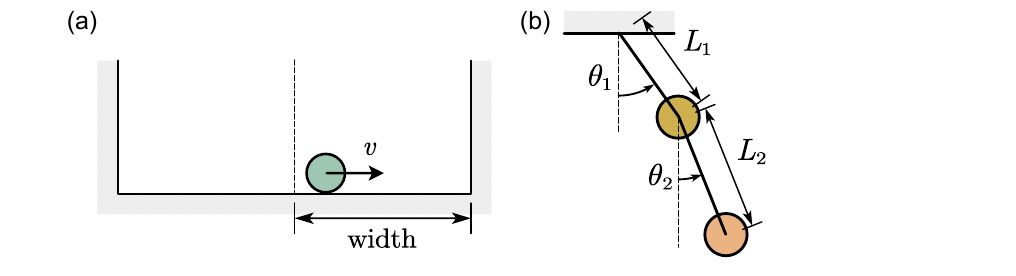}
    \caption{(color online). (a) The wall bouncing system. (b) The double pendulum system.}
    \label{fig:Other}
\end{figure}

\subsection{Wall bouncing system}
The governing equation for the wall bouncing system (Fig.~\ref{fig:Other}(a)) is
\begin{align}
    \ddot{x} + &F(x) = 0 \quad s.t. \quad x(0) = x_0, \ \dot{x}(0) = \dot{x}_0 \nonumber \\
    &F(x) = \Biggl\{ 
    \begin{array}{l}
         -k(x-w), \ x \geq w\\
         0 \qquad \qquad , \ |x| < w \\
         -k(x+w), \ x \leq -w
    \end{array}
\end{align}
$x$ and $v$ are the particle position and velocity, $k=1000$ N/m is a large constant to approximate a stiff wall, $w$ is the (half-)width of the potential well, as shown in Fig.~\ref{fig:Other}(a). 
The system has a potential energy well with the following form
\begin{equation}
    E(x) = \left\{
    \begin{aligned}
    &0, &|x| < w\\
    &\infty, &|x| \geq w
    \end{aligned}
    \right.
\end{equation}
The training dataset contains 10 system instances, with the width increasing from 0.1 m to 1.0 m by 0.1 m. Trajectories of multiple initial conditions (initial position 0 m, and initial velocities ranging from 0.1 m/s to 1.0 m/s) with stepsize 10 ms and time span 10 s are generated for each instance. During training, a batch of 100 randomly pulled-out trajectories of 1 s is used for each epoch.

\begin{figure}[htbp]
    \centering
    \includegraphics[width=0.46\textwidth]{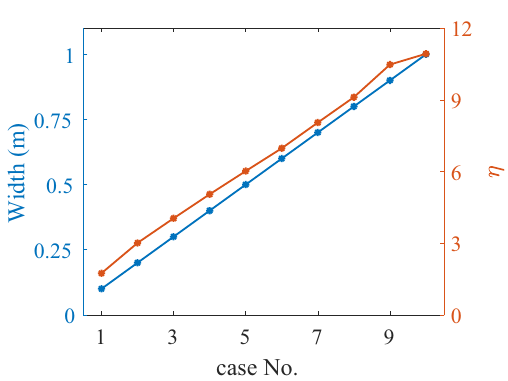}
    \caption{(color online). The true widths of the wall bouncing system and the learnt adaptation parameters $\eta$ are in perfect correlation (99.87\%).}
    \label{fig:Gap}
\end{figure}

In this case the intermediate output of the NN is the energy and the force is derived by taking the derivative of the output with respect to the input, i.e.
\begin{equation}
    \label{Eq:Energy}
    F = \frac{\partial E}{\partial x} = \frac{\partial (\mathrm{NN}_{\boldsymbol{\theta}}(x;\eta)+\mathrm{NN}_{\boldsymbol{\theta}} (-x;\eta))}{\partial x}
\end{equation}
The second equality takes advantage of the assumption that the energy is symmetric with respect to $x$. The learning results show $\eta \in \mathbb{R}$ to be in perfect correlation with the width $w$ of the potential well (Fig.~\ref{fig:Gap}). In other words, we can control the width of the constructed potential well of the NN, which is another way to interpret the physical meaning of the adaptation parameter $\eta$.
\begin{figure}[htbp]
    \centering
    \includegraphics[width=0.85\textwidth]{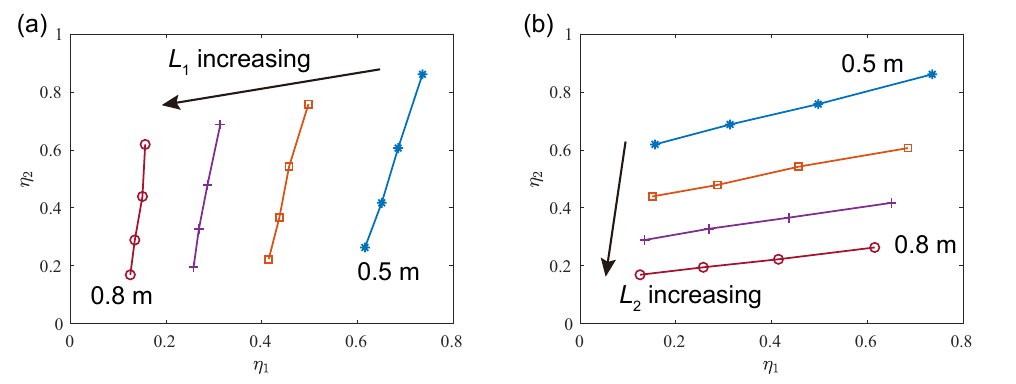}
    \caption{(color online). The learnt latent space of adaptation parameters for the double pendulum system. (a) Each marked line shows systems with the same $L_1$. (b) Each marked line shows systems with the same $L_2$. There are clearly two directions in the latent space (indicated by the arrows) corresponding to the change of physical parameters $L_1$ and $L_2$.}
    \label{fig:DoublePendulum}
\end{figure}

\subsection{Double pendulum}
The double pendulum, as shown in Fig.~\ref{fig:Other}(b), has two masses $m_1=m_2=1$ kg and arm lengths $L_1$ and $L_2$. The governing equations are
\begin{equation}
    \begin{gathered}
    \dot{\theta}_1=\omega_1 \\
    \dot{\theta}_2=\omega_2 \\
    \dot{\omega}_1=\frac{-g\left(2 m_1+m_2\right) \sin \theta_1-m_2 g \sin \left(\theta_1-2 \theta_2\right)-2 \sin \left(\theta_1-\theta_2\right) m_2\left(\omega_2^2 L_2+\omega_1^2 L_1 \cos \left(\theta_1-\theta_2\right)\right)}{L_1\left(2 m_1+m_2-m_2 \cos \left(2 \theta_1-2 \theta_2\right)\right)} \\
    \dot{\omega}_2=\frac{2 \sin \left(\theta_1-\theta_2\right)\left(\omega_1^2 L_1\left(m_1+m_2\right)+g\left(m_1+m_2\right) \cos \theta_1+\omega_2^2 L_2 m_2 \cos \left(\theta_1-\theta_2\right)\right)}{L_2\left(2 m_1+m_2-m_2 \cos \left(2 \theta_1-2 \theta_2\right)\right)}
    \end{gathered}
\end{equation}
The training dataset contains 16 system instances, a mesh of $L_1=[0.5,0.6,0.7,0.8]$ m and $L_2=[0.5,0.6,0.7,0.8]$ m. Trajectories of initial locations [$\pi/4$,$\pi/4$] and initial velocities [0,0] with stepsize 10 ms and time span 10 s are generated for each system. During training, a batch of 100 randomly pulled-out trajectories of 1 s is used for each epoch. Task adaptation takes 5 steps. The learnt latent space of adaptation parameters is shown in Fig.~\ref{fig:DoublePendulum}. It is clear that two directions exist corresponding to the variation of physical parameters $L_1$ and $L_2$. This again underlines the interpretability of $\boldsymbol{\eta} \in \mathbb{R}^2$.

% \section{Different initialization as one system}
\section{Comparison between iMODE and training from scratch}
\begin{figure}
    \centering
    \includegraphics[width=0.85\textwidth]{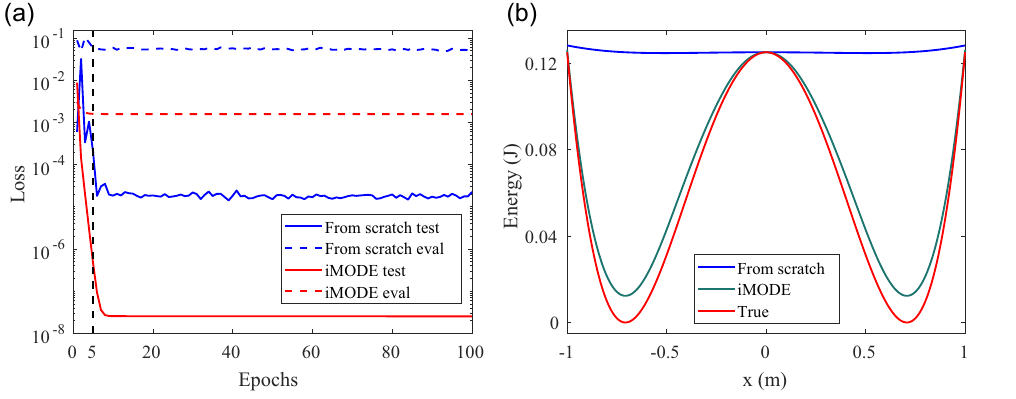}
    \caption{(color online). (a) The training/adaptation and evaluation performance of the TFS and iMODE NNs, given a single trajectory of the bistable system with physical parameters $k_1=-1.0$ and $k_3=2.0$, and initial condition $x_0=0.7$ m and $\dot{x}_0=0$ m/s. (b) The iMODE learns the correct double-well potential energy function while the TFS approach learns nothing due to data scarcity.}
    \label{fig:Special}
\end{figure}
We compare the performance of iMODE adaptation to the ``training from scratch'' (TFS) approach. The iMODE adaptation starts with a weight initialization trained from a training dataset. It updates the adaptation parameter $\boldsymbol{\eta}\in\mathbb{R}^2$ on a testing dataset, which is not included in the training dataset. The TFS approach uses the same NN architecture and hyperparameters as in the iMODE. The TFS NN is randomly initialized and all the weights are updated on the same testing dataset. After training the TFS NN and adapting the iMODE $\boldsymbol{\eta}$ NN using the same testing dataset, the two NNs are evaluated on an unseen evaluation dataset. As shown in Fig.~3(a), the iMODE significantly outperforms the TFS approach in terms of adaptation speed (v.s. training speed in the TFS approach) and evaluation accuracy. This means that the iMODE approach can learn the dynamics of an unseen system more rapidly and predict future events more accurately than a TFS NN. This observation is pronounced in the following case: we feed these two NNs a single trajectory of the bistable system with physical parameters $k_1=-1.0$ and $k_3=2.0$, and initial condition $x_0=0.7$ m and $\dot{x}_0=0$ m/s. After training/adaptation, we evaluate the TFS and iMODE NNs on trajectories of the same system but with differential initial conditions. The training/adaptation and evaluation curves are shown in Fig.~\ref{fig:Special}(a). The iMODE outperforms the TFS approach in both training/adaptation and evaluation accuracy. The learnt energy functions of both approaches are compared in Fig.~\ref{fig:Special}(b). The energy function of the TFS NN is totally incorrect due to the data scarcity. Under this specific initial condition, the bistable system is only oscillating intra-well. So the information contained in the trajectory is insufficient to depict the entire potential energy surface. Meanwhile the iMODE NN learns an accurate double-potential-well function from the same data because appropriate prior knowledge on the energy functions of bistable systems (i.e. double-well) is already embedded in its weight initialization.

\section{Dimension determination of physical parameters with PCA \label{supp:pca}}
The workflow of using PCA to determine the optimal dimension $d$ of the adaptation parameters $\boldsymbol{\eta}$ for a parametric system is: (1) Make a rough guess $\tilde{d}$ on the dimension, then run the iMODE algorithm on the trajectories of $N_s$ systems; (2) Form a matrix with the results $M=[\boldsymbol{\eta}_1,\boldsymbol{\eta}_2,\dots,\boldsymbol{\eta}_{N_s}]$; (3) Perform PCA on $M$. A significant portion of variance (e.g. 99\%) will be preserved in the first $\hat{d}$ dimensions, an estimation for $d$; (4) repeat the process with different initial guesses $\tilde{d}$. The optimal dimension is more credible when different $\tilde{d}$ results in the same $\hat{d}$.

\begin{figure}[htbp]
    \centering
    \includegraphics[width=0.95\textwidth]{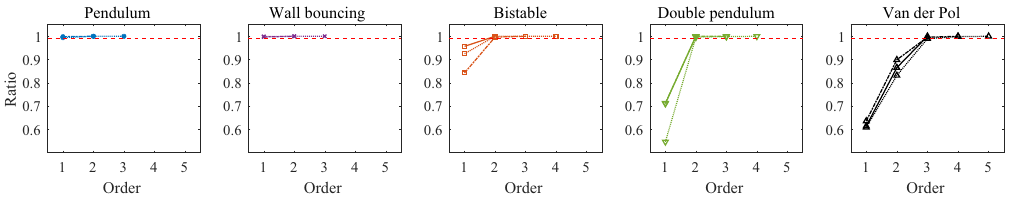}
    \caption{(color online). PCA can be used to determine the optimal dimensions of the adaptation parameters $\boldsymbol{\eta}$ in the studied systems. These optimal dimensions prove to equal the true dimension $d_{\boldsymbol{\phi}}$ of the physical parameters $\boldsymbol{\phi}$ of the systems. The red dashed line indicates 0.99.}
    \label{fig:PCA}
\end{figure}
The PCA determination results of all systems are shown in Fig.~\ref{fig:PCA}. The red dashed lines mark the $99\%$ variance preservation. Dotted lines in each case mean that the initial guess $\tilde{d}$ is the dimension of real physical parameters $d_{\boldsymbol{\phi}}$ plus 2. For example, in the Van der Pol system, the dotted line means that the initial guess $\tilde{d}=5$. After the PCA, if we preserve 4 or 3 principal components, the variance energy is still preserved by more than 99\%. If we further reduce the number of preserved principal components to 2 or 1, we see a sudden drop (to below the 99\% threshold), which indicates the optimal dimension to be 3. The $\tilde{d}$ for dashed-dotted and solid lines are the dimension of real physical parameters $d_{\boldsymbol{\phi}}$ plus 1 and 0 respectively. With different $\tilde{d}$, we can repeatedly confirm the optimal dimension of $\boldsymbol{\eta}$, to be 3 in the case of Van der Pol system (which is the true dimension of physical parameters). For other systems, the workflow is the same.

\section{\emph{Neural Gauge} diffeomorphism}
As suggested by the PCA analysis in Section~\hyperref[supp:pca]{S\ref*{supp:pca}}, the adaptation parameters $\{\vec{\eta}_i\}_{i=1}^{N_{\mathrm{s}}}$ that are adapted to the system instances belonging to a family of dynamical systems occupy a $d_{\vec{\phi}}$-dimensional manifold, even if the latent space they reside in is $d_{\vec{\eta}}$-dimensional and $d_{\vec{\eta}} \geq d_{\vec{\phi}}$. Therefore, a diffeomorphism can be established mapping $\vec{\eta}$ in the latent space to their corresponding physical parameters ($\{\vec{\phi}_i\}_{i=1}^{N_{\mathrm{s}}}$) even if their dimensions do not match, assuming $d_{\vec{\eta}} \geq d_{\vec{\phi}}$. Practically, the neural ODE modelling such diffeomorphism can be defined as ${\mathrm{d}\vec{z}(t)} / {\mathrm{d} t} = \vec{g}_{\vec{\xi}}(\mathbf{z})$, such that for $i=1,\ldots,N_s$, starting from a given point in the latent space, $\vec{z}(0) = \vec{\eta}_i$, the state $\vec{z}$ at $t=1$, $\vec{z}(1) = \begin{bmatrix}\vec{\phi}_i^{\mathrm{T}} & 0 & \ldots & 0 \end{bmatrix}^{\mathrm{T}}$ is the concatenation of corresponding physical parameters and $d_{\vec{\eta}} - d_{\vec{\phi}}$ padding zeros.

\section{Complex cases}
%Invariance and induced equivariance on the Slinky case.
\subsection{Slinky: the Euclidean symmetric neural network}

% In the Slinky case, we approximate the Slinky dynamics in 2D and embed physical symmetries into the iMODE NN. The NN is constructed in a specific form so that the output energy is invariant to Euclidean transformations of the Slinky coordinates, including translation (tr), rotation (ro), and reflection (re), i.e.,
% \begin{align}
% \energy = \neuralfun_{\boldsymbol{\theta}}(\mathbf{x};\boldsymbol{\eta})&=\neuralfun_{\boldsymbol{\theta}}\left(tr(\mathbf{x});\boldsymbol{\eta}\right)=\neuralfun_{\boldsymbol{\theta}}\left(ro(\mathbf{x});\boldsymbol{\eta}\right) \nonumber \\
% &=\neuralfun_{\boldsymbol{\theta}}\left(re(\mathbf{x});\boldsymbol{\eta}\right)
% \end{align}
% The force $\force$ is derived by Eq.~\eqref{Eq:Energy} and is proved to be equivariant to Euclidean transformations of the Slinky coordinates, regardless of $\boldsymbol{\eta}$ (refer to \cite{Li2022Rapidly} for detailed analysis).

\begin{figure}[htbp]
    \centering
    \includegraphics[width=0.65\textwidth]{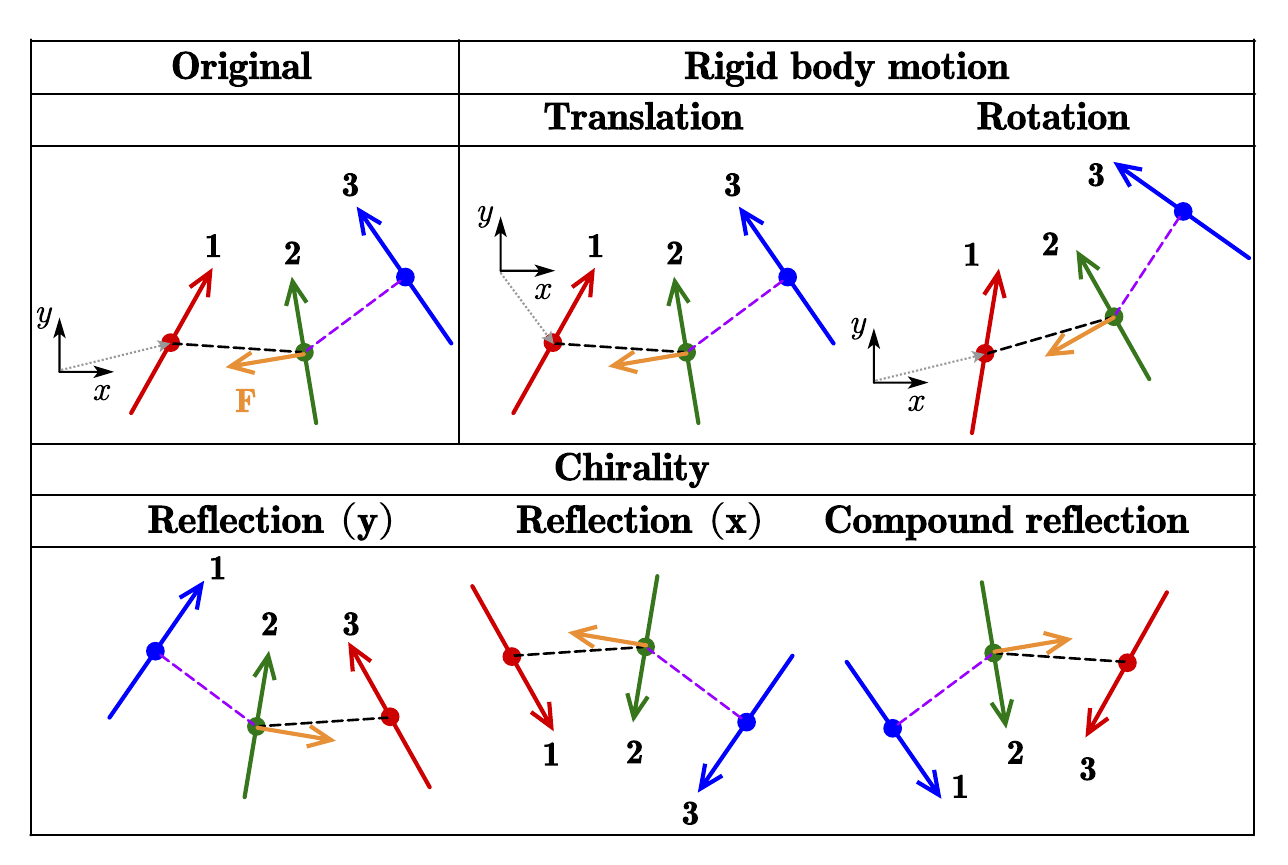}
    \caption{(color online). The Euclidean invariance on energy and induced equivariance on force of the NN used in the Slinky system case. All the Euclidean transformed configurations have the same energy as the original configuration. The elastic forces on the middle bars are Euclidean-transformed accordingly.}
    \label{fig:Invariance}
\end{figure}
The NN used in the Slinky case follows the Euclidean symmetric neural network (ESNN) \cite{Li2022Rapidly} architecture.
The Slinky is decomposed into 40 consecutive triplets, i.e., the 2D representation of 3 adjacent cycles. We denote the coordinates of the $i$th triplet as $\vec{\xi}_i=\left[\mathbf{x}_{i-1}^{\mathrm{T}}, \mathbf{x}_i^{\mathrm{T}}, \mathbf{x}_{i+1}^{\mathrm{T}}\right]^{\mathrm{T}} \in \mathbb{R}^9$. $\mathbf{x}_i \in \mathbb{R}^3$ is the coordinates of the $i$th bar, including the $x$ and $y$ coordinates of the bar center and the inclination angle of the bar. The potential energy associated with the middle bar of a triplet is only a function of the coordinates of the 3 bars of the same triplet (and of the adaptation parameters), i.e.
\begin{equation}
    E_i = E_i(\vec{\xi}_i;\boldsymbol{\eta})
\end{equation}
In the following we will omit the subscript $i$ for brevity. The induced force $\mathbf{F}$ from $E$ is
\begin{equation}
\label{Eq:SlinkyForce}
    \mathbf{F} = \frac{\partial E}{\partial \mathbf{x}_i} = \frac{\partial E(\mathbf{z};\boldsymbol{\eta})}{\partial \mathbf{x}_i} = \frac{\partial E(\vec{\xi};\boldsymbol{\eta})}{\partial \mathbf{x}_i}
\end{equation}
where $\mathbf{z} \in \mathbb{R}^6$ is the relative coordinates between the bars of the $i$th triplet. We enforce Euclidean invariance, i.e., translational, rotational, and chiral invariance, on $E$ with respect to $\vec{\xi}$, by taking $E(\mathbf{z})$ the following form, i.e., the ESNN 
\begin{align}
E(\mathbf{z};\boldsymbol{\eta}) &= \mathrm{NN}_{\boldsymbol{\theta}}(\mathbf{z};\boldsymbol{\eta})+\mathrm{NN}_{\boldsymbol{\theta}}\left(R_x(\mathbf{z});\boldsymbol{\eta}\right)+\mathrm{NN}_{\boldsymbol{\theta}}\left(R_y(\mathbf{z});\boldsymbol{\eta}\right)+\mathrm{NN}_{\boldsymbol{\theta}}\left(R_x\left(R_y(\mathbf{z})\right);\boldsymbol{\eta}\right)
\end{align}
where $R_x(\cdot)$ and $R_y(\cdot)$ denote reflection with respect to the $x$ and $y$ axes. Note that
\begin{align}
R_x\left(R_x(\cdot)\right)=I(\cdot), \  &R_y\left(R_y(\cdot)\right)=I(\cdot), \ \text{and } R_x\left(R_y(\cdot)\right)=R_y\left(R_x(\cdot)\right)
\end{align}
where $I(\cdot)$ is the identity operation. It is easy to prove the chiral invariance of $E$, i.e.,
\begin{align}
E(\mathbf{z};\boldsymbol{\eta})&=E\left(R_x(\mathbf{z});\boldsymbol{\eta}\right)=E\left(R_y(\mathbf{z});\boldsymbol{\eta}\right) =E\left(R_x\left(R_y(\mathbf{z})\right);\boldsymbol{\eta}\right)
\end{align}
Then from Eq.~\eqref{Eq:SlinkyForce}, $\mathbf{F}$ is equivariant to rigid body and chiral transformations on $\vec{\xi}$, as shown in Fig.~\ref{fig:Invariance}, including translation, rotation, and reflection, regardless of $\boldsymbol{\eta}$. The ESNN will be applied on each triplet in the Slinky to calculate the elastic force acting on each bar. The assembled force vector is used to update the system state inside the NODE framework. The difference between the true and predicted trajectories is used to update the ESNN weights $\boldsymbol{\theta}$. After training and performing trajectory predictions in 2D, a geometric method can be used to reconstruct the 3D Slinky configurations \cite{Li2022Rapidly}. See \cite{Li2022Rapidly} for more implementation details.

\begin{figure}
    \centering
    \includegraphics[width=0.90\textwidth]{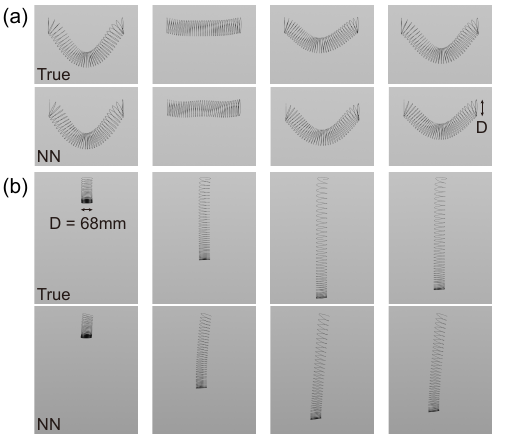}
    \caption{(color online). (a) The testing performance of the iMODE model on an unseen Slinky (of an unseen Young's modulus) with the same boundary condition as the training dataset. Top and bottom rows are ground truth and the iMODE model prediction at 0.28, 0.47, 0.65, 0.83 s (left to right). (b) The testing performance of the iMODE model on unseen initial and boundary conditions. Top and bottom rows are ground truth and the iMODE model prediction at 0.15, 0.32, 0.48, 0.65 s (left to right).}
    \label{fig:Slinky}
\end{figure}

The training dataset contains 4 Slinkies of different Young's modulus (50, 60, 70, and 80 GPa). The Slinkies are clamped at both ends and freely drop under gravity from a horizontal initial configuration. Two inner steps are taken to update $\eta \in \mathbb{R}$ for each Slinky. Note when $\eta$ is updated, the NN always preserves energy invariance and force equivariance with respect to the coordinates of the Slinky. After training the NN, we perform task adaptation (2 steps) on a unseen Slinky of Young's modulus 56 GPa and observe a good fitting result (Fig.~\ref{fig:Slinky}(a)). The resulting NN is then directly applied to computation under an unseen boundary condition and Slinky orientation (the bottom end is free and the Slinky drops under gravity from a vertical initial configuration) without any modification (Fig.~\ref{fig:Slinky}(b)). We can achieve this because the model-agnostic nature of the iMODE method allows us to embed the Euclidean symmetries into the NN.

\begin{figure}[htbp]
    \centering
    \includegraphics[width=0.85\textwidth]{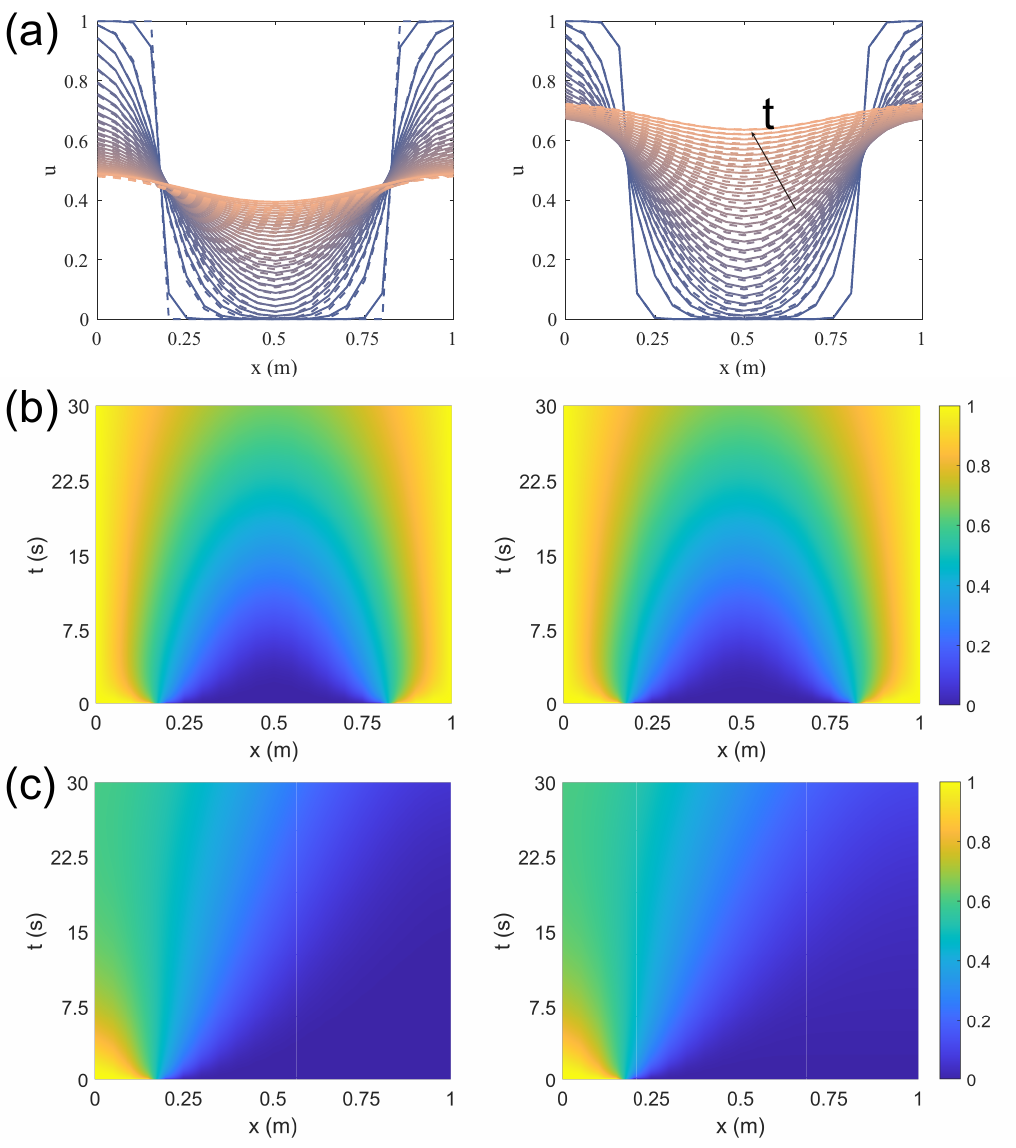}
    \caption{(color online). (a) The training results of the iMODE algorithm on the KPP system for $r=0.01$ (left) and $r=0.05$ (right). Solid lines are ground truth. Dashed lines are iMODE predictions. The arrow indicates time marching of $u$. (b) The iMODE testing results on an unseen system $r=0.034$ with an unseen boundary condition. The ground truth (left) match the iMODE prediction (right) well. (c) The iMODE testing results ($r=0.034$) on an unseen initial condition from the training dataset. The ground truth (left) match the iMODE prediction (right) well.}
    \label{fig:KPP}
\end{figure}

\subsection{Kolmogorov-Petrovsky-Piskunov (KPP) equation}

To solve the KPP equation, we discretize the spatial domain [0,1] into 20 segments. So the the partial differential equation system (here $x$ denotes the spatial coordinate)
\begin{equation}
\frac{\partial u}{\partial t}=D \frac{\partial^2 u}{\partial x^2}+r u(1-u)
\end{equation}
is represented by an ordinary differential equation system containing 21 variables. The diffusion term is approximated by 2nd-order central difference and the diffusivity is assumed known. The meta-learning is performed to learn the reaction term with different reaction strength coefficients $r$ (without knowing the mathematical form). The training dataset contains 5 systems with $r=0.01,\ 0.02,\ 0.03,\ 0.04,\ 0.05$. The Neumann boundary condition $u^{\prime}(0)=u^{\prime}(1)=0$ ($\prime$ denotes derivative with respect to $x$) is used across the training dataset. The iMODE task adaptation takes 5 iterations. The training results for $r=0.01,\ 0.05$ are shown in Fig.~\ref{fig:KPP}(a). The iMODE NN is then adapted on the data from a unseen system instance with $r=0.034$. The resulted NN is directly applied to computation with unseen initial and boundary conditions (Dirichlet type $u(0)=u(1)=1$). The results of the latter are shown in Fig.~\ref{fig:KPP}(b) and a good agreement is observed. This again validates the capability of the iMODE algorithm to fast adapt on unseen complex parametric systems and accurately predict on initial and boundary conditions different from those in the training dataset.

Another testing result for the KPP system is shown in Fig.~\ref{fig:KPP}(c). The testing has the same type of boundary condition ($u^{\prime}(0)=u^{\prime}(1)=0$) as the training dataset but an unseen initial condition. The prediction (right) matches the ground truth (left) well.
% \section{Discussion}

% \subsection{Irregular data mesh}

\begin{figure}[htbp]
    \centering
    \includegraphics[width=0.90\textwidth]{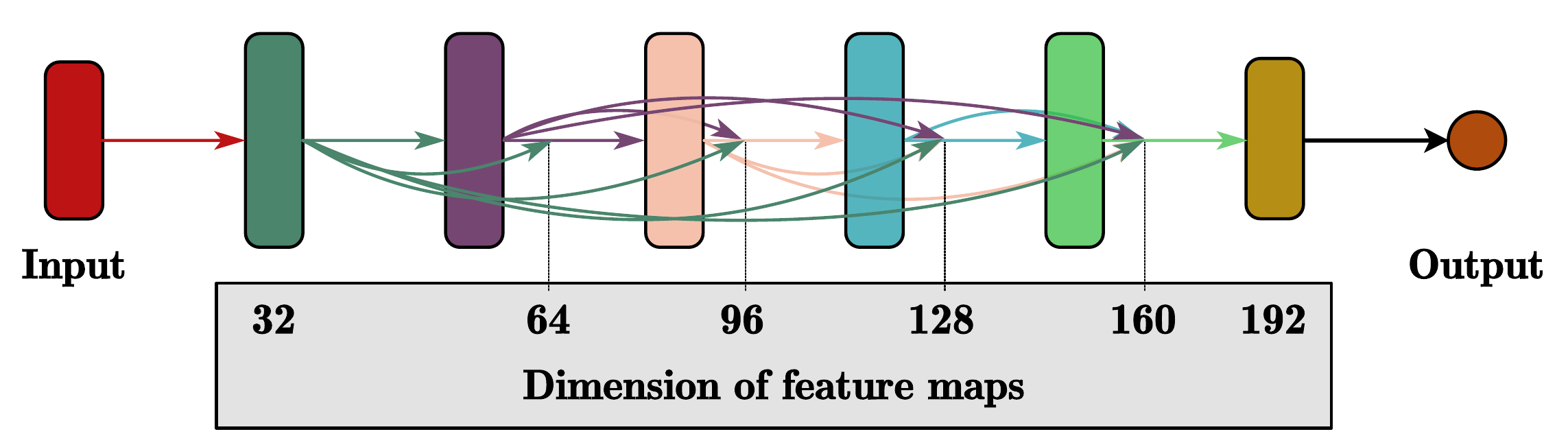}
    \caption{(color online). The DenseNet-like structure.}
    \label{fig:NN}
\end{figure}

\section{Neural network architecture}
Throughout this letter, we use a DenseNet-like architecture \cite{Huang2017Densely} for our neural networks (NNs), where shortcut pathways are created for a layer from all its previous layers. It takes in the input and first increases the feature dimension to 32 by a fully-connected (FC) layer. Then the feature is passed through FC layers with Softplus activation. A new feature with an increased dimension is formed by concatenating the previous feature with the FC layer output, i.e.,
\begin{align}
    &\mathbf{f}_i = 
    \begin{bmatrix}
    \textrm{FC}(\mathbf{f}_{i-1})  \in \mathbb{R}^{32} \\
    \mathbf{f}_{i-1}
    \end{bmatrix}, \quad i=(1,2,\dots,5) \nonumber \\
    &\mathbf{f}_0 = \textrm{FC}(\mathrm{Input}) \in \mathbb{R}^{32} \nonumber \\
    &\mathrm{Output} = \textrm{FC}(\mathbf{f}_{5})
\end{align}
where $\mathbf{f}_i$ is the feature map for the $i$th layer. After passing through 5 densely connected layers, the feature dimension is increased to 192. This feature is then passed through a FC layer with no activation to produce the final output.

For pendulum, bistable, wall bouncing, and Slinky systems, the NN input is the vector concatenating the system position $\mathbf{x}$ and the adaptation parameter $\boldsymbol{\eta}$, i.e. $[\mathbf{x}^{\mathrm{T}}, \boldsymbol{\eta}^{\mathrm{T}}]^{\mathrm{T}}$. The output is a scalar, i.e. the energy of the system. The force vector is calculated by back-propagating the NN output with respect to $\mathbf{x}$. For Van der Pol system, the input is the vector concatenating the system state $\mathbf{y}$ and $\boldsymbol{\eta}$, i.e. $[\mathbf{y}^{\mathrm{T}}, \boldsymbol{\eta}^{\mathrm{T}}]^{\mathrm{T}}$. The output is the force vector. For KPP system, the input is $[u, \eta]$. The output is the reaction forcing term.

\section{Supplementary movie}

\noindent \textbf{Movie S1.} The diffeomorphism for the bistable system. The data points are transformed from the physical space (subtracted mean) to the latent space of adaptation parameters (subtracted mean). The right subplot is the enlarged view of the left plot. \\

\noindent \textbf{Movie S2.} The diffeomorphism for the Van der Pol system. The data points are transformed from the physical space (subtracted mean) to the latent space of adaptation parameters (subtracted mean). The right subplot is the enlarged view of the left plot.

\end{document}